\documentclass[10pt,conference,a4paper]{IEEEtran}

\usepackage{times}
\usepackage{epsfig}
\usepackage{graphicx}
\usepackage{amsmath}
\usepackage{amssymb}
\usepackage{subfiles}
\usepackage{pifont}
\usepackage{cite}
\usepackage{booktabs}
\usepackage{multirow}
\usepackage{xcolor}

\usepackage[pagebackref=true,breaklinks=true,letterpaper=true,colorlinks,bookmarks=false]{hyperref}

\graphicspath{{./images/}}

\begin{document}

\title{Camera Bias in a Fine Grained Classification Task}

\author{Philip T. Jackson$^1$ \quad\quad Stephen Bonner$^1$ \quad\quad Ning Jia$^1$ \quad\quad Christopher Holder$^1$ \\ \quad\quad Jon Stonehouse$^2$ \quad\quad Boguslaw Obara$^{1}$\\
$^{1}$Department of Computer Science, Durham University, Durham, UK\\
$^{2}$Procter and Gamble, Reading, UK\\
{\tt\small \{philip.jackson,stephen.bonner,ning.jia,c.j.holder,boguslaw.obara\}@durham.ac.uk}\\
{\tt\small stonehouse.jr@pg.com}}

\maketitle

\begin{abstract}
   We show that correlations between the camera used to acquire an image and the class label of that image can be exploited by convolutional neural networks (CNN), resulting in a model that ``cheats'' at an image classification task by recognizing which camera took the image and inferring the class label from the camera. We show that models trained on a dataset with camera / label correlations do not generalize well to images in which those correlations are absent, nor to images from unencountered cameras. Furthermore, we investigate which visual features they are exploiting for camera recognition. Our experiments present evidence against the importance of global color statistics, lens deformation and chromatic aberration, and in favor of high frequency features, which may be introduced by image processing algorithms built into the cameras.
\end{abstract}

\section{Introduction}
\label{sec:intro}

Convolutional neural networks (CNN) sometimes learn to satisfy their objective functions in ways we do not intend, typically by exploiting some subtle idiosyncrasy in the training data. For example, in \cite{fong2017interpretable} a CNN trained on ImageNet was found to be recognizing chocolate sauce pots by the presence of a spoon, because many of the chocolate sauce pots in the ImageNet dataset are indeed accompanied by a silver spoon. While effective at minimizing the loss function at training time, these clever exploits usually result in the model becoming brittle, as it is relying on characteristics that are specific to the training set and are not representative of the wider world. This tends to manifest as domain bias, whereby the model fails to generalise well to instances from other datasets with different idiosyncrasies.

We investigate a real world applied computer vision problem in which severe domain bias was caused by strong correlations between camera model and class label. Since the training dataset consists of two classes acquired with different cameras, the model learns to predict the class label by recognizing the camera that captured the image. Since the sets of cameras used to acquire the two classes are non intersecting, this is sufficient to achieve perfect training accuracy, whilst learning nothing about the task itself. Our task has characteristics typical of industrial deep vision projects, and we believe the lessons learned will be useful to many deep learning practitioners working on similar projects. By illuminating the sometimes counterintuitive means by which CNNs can classify images, our work is also relevant to the ongoing quest for algorithmic transparency and accountability in machine learning.

The task itself is to discriminate between shampoo bottles from two different manufacturers, which are distinguished only by very small differences in the printing of a batch code on the underside of the bottle. These differences are caused by different industrial printers being used, are independent of the actual character string that is printed, and are subtle enough that detecting them by eye is difficult even for trained experts. This therefore constitutes a fine grained binary classification problem, in which the intra-class variance is high relative to the inter-class variance. 

Fine grained classification is difficult, so one might intuitively expect a model to cheat more often on such tasks, if the correct decision function is more complex compared to a cheating rule. On the other hand, in this instance the exploit of recognizing cameras is also a fine grained classification task, and in general it is not obvious which tasks are ``harder'' for a CNN to learn. CNNs have been known to cheat by detecting patterns which are barely perceptible to humans, such as chromatic aberration \cite{doersch2015unsupervised}. CycleGAN even cheats its reconstruction loss by inserting steganographic codes into its converted images, which it then uses to reconstruct the originals \cite{chu2017cyclegan}. 

In this paper, we closely examine an instance of a model cheating on a real world visual classification task, and attempt to answer the following questions:

\begin{enumerate}
    \item Is it possible for a CNN to recognize camera types when explicitly trained to do so?
    \item Can we prove that the same CNN cheats on the task of manufacturer classification by recognizing camera types?
    \item Does the propensity toward cheating depend on model architecture?
    \item How exactly does a CNN recognize camera types?
\end{enumerate}

Section~\ref{sec:lit_review} reviews relevant literature in fine grained classification and overfitting, while Section~\ref{sec:dataset} describes out dataset and classification task in detail. Section~\ref{sec:experiments} investigates the above questions systematically with a series of experiments, and in Section~\ref{sec:conclusion} we discuss our findings and draw conclusions.

\section{Previous Work}
\label{sec:lit_review}


Two major branches of literature are relevant to our work: source camera identification from images, and understanding deep neural networks.

\subsection{Camera / Image Sensor Pattern Identification}

Because our work concerns accidental camera detection, a brief review of deliberate camera detection methods is warranted, as it may shed some light on how our model learns to cheat. Many techniques have been developed to trace digital photos back to their camera of origin, primarily by the digital forensics community \cite{fridrich2009digital}. Such techniques can be used to detect doctored images or videos, where images or frames from different cameras are spliced together \cite{cozzolino2019extracting,cozzolino2019noiseprint}. Most of these methods revolve around extracting a unique sensor noise fingerprint from the image, and matching it against the reference patterns of known cameras. Since sensor noise is a complex phenomenon with multiple sources (e.g. photonic noise, lens imperfections, dust particles, dark currents, non-uniform pixel sensitivity), there are many ways of doing this. Geradts et al. \cite{geradts2001methods} identify cameras by their unique patterns of dead and hot pixels, however not all cameras have dead pixels, and some remove them via post-processing. Kharrazi et al. \cite{kharrazi2004blind} train an SVM to recognize five different cameras based on hand-engineered feature vectors extracted from images. This approach achieves up to $95\%$ classification accuracy, but this is too low for forensic purposes. Choi et al. \cite{choi2006source} take a similar SVM based approach, additionally showing that radial lens distortion is a useful feature for identifying cameras. Unlike noise based approaches, lens distortion can identify models of camera but not individuals. Kurosawa et al. \cite{kurosawa1999ccd} recognizes cameras by dark current noise, which is a small, constant signal emitted by a CCD, varying randomly from pixel to pixel. Although every digital camera has such a noise pattern and it will always be unique, it can only be acquired from dark frames where no light strikes the sensor, and is only a small component of sensor noise. Lukas et al. \cite{lukavs2006digital} propose a more robust method that exploits the non-uniform sensitivity to light among sensor pixels, which is a much stronger component and does not require dark frames to measure. 

Another feature of consumer cameras that has thwarted a previous deep learning experiment \cite{doersch2015unsupervised} is chromatic aberration, in which different wavelengths of light are refracted by different amounts by the lens. This results in colored fringes around the edges of objects. This too has been used in digital forensics \cite{johnson2006exposing}. 


Recently, CNN-based methods have shown great potential in digital camera identification from images using standard supervised training \cite{tuama2016camera,bondi2017preliminary,yao2018graphics}, proving that CNNs are indeed able to infer which camera acquired a digital image.

\subsection{Understanding Deep Convolutional Neural Networks}

CNNs are often seen as something of a black box, with no clear consensus as to what information they are using to reach their decisions, how that information is represented internally, or what are the specific roles of their individual components. Attempts to answer these questions can be divided into two strands, feature visualization and attribution.

Feature visualization aims to clarify the function of neurons or channels, by synthesizing images that maximize their activation \cite{olah2017feature}. Simonyan et al. \cite{simonyan2013deep} investigate what patterns CNNs look for in each image class by performing gradient ascent in image space, to maximize the activation of an output class neuron. Yosinski et al. \cite{yosinski2015understanding} do the same but with better regularization, producing more natural looking images. Mahendran et al. \cite{mahendran2015understanding} treat intermediate CNN representations as functions which they can invert via gradient ascent in image space. This yields images that the CNN maps to the same representation as the original image, implying that they ``look the same'' to the CNN. Nguyen et al. \cite{nguyen2016synthesizing} find natural looking images that maximally activate feature maps by searching the manifold learned by a generative adversarial network, rather than the full image space. Fong et al. \cite{fong2018net2vec} show evidence that far from feature maps learning separate, well defined concepts, the relationship between feature maps and semantic concepts is many-to-many, with each feature map involved in the detection of several concepts and most concepts activating multiple feature maps.

Attribution investigates which parts of an image contribute most to a CNN's decision - often expressed as ``where the model is looking''. Zeiler and Fergus \cite{zeiler2014visualizing} propose two methods to this end: occlusion mapping, in which the importance of an image patch is measured as the reduction in class probability when it is obscured, and backpropagation of class probability gradients into image pixels. Both of these methods yield saliency maps showing which parts of the image have the greatest effect on the output when changed, corresponding to the notion of how much they contributed to the network's decision. Another popular approach is guided backpropagation \cite{springenberg2014striving}, which refines the gradient saliency maps of \cite{zeiler2014visualizing} by zeroing out negative gradients at every backpropagation step, so as to focus only on image parts that contribute positively to a particular class. A much faster alternative to occlusion mapping (which must run a forward pass for each test patch) is class activation mapping (CAM) \cite{zhou2016learning}, which uses final layer feature maps as saliency maps, weighted and summed according to the weight of their connection to the class neuron in question. This approach requires that the output layer takes its input directly from mean pooled feature maps (as is the case with GoogLeNet and ResNet but not for networks with fully connected layers such as AlexNet). Selvaraju et al. \cite{selvaraju2017grad} address this by using mean pooled gradients as a proxy for direct connection weights, allowing feature maps from any layer in any network to be used as saliency maps. Another technique by Fong et al. \cite{fong2017interpretable} learns a mask that causes a model to misclassify an image while obscuring the smallest area possible. 
\section{Dataset}
\label{sec:dataset}

\begin{figure}
    \centering
    \includegraphics[width=\linewidth]{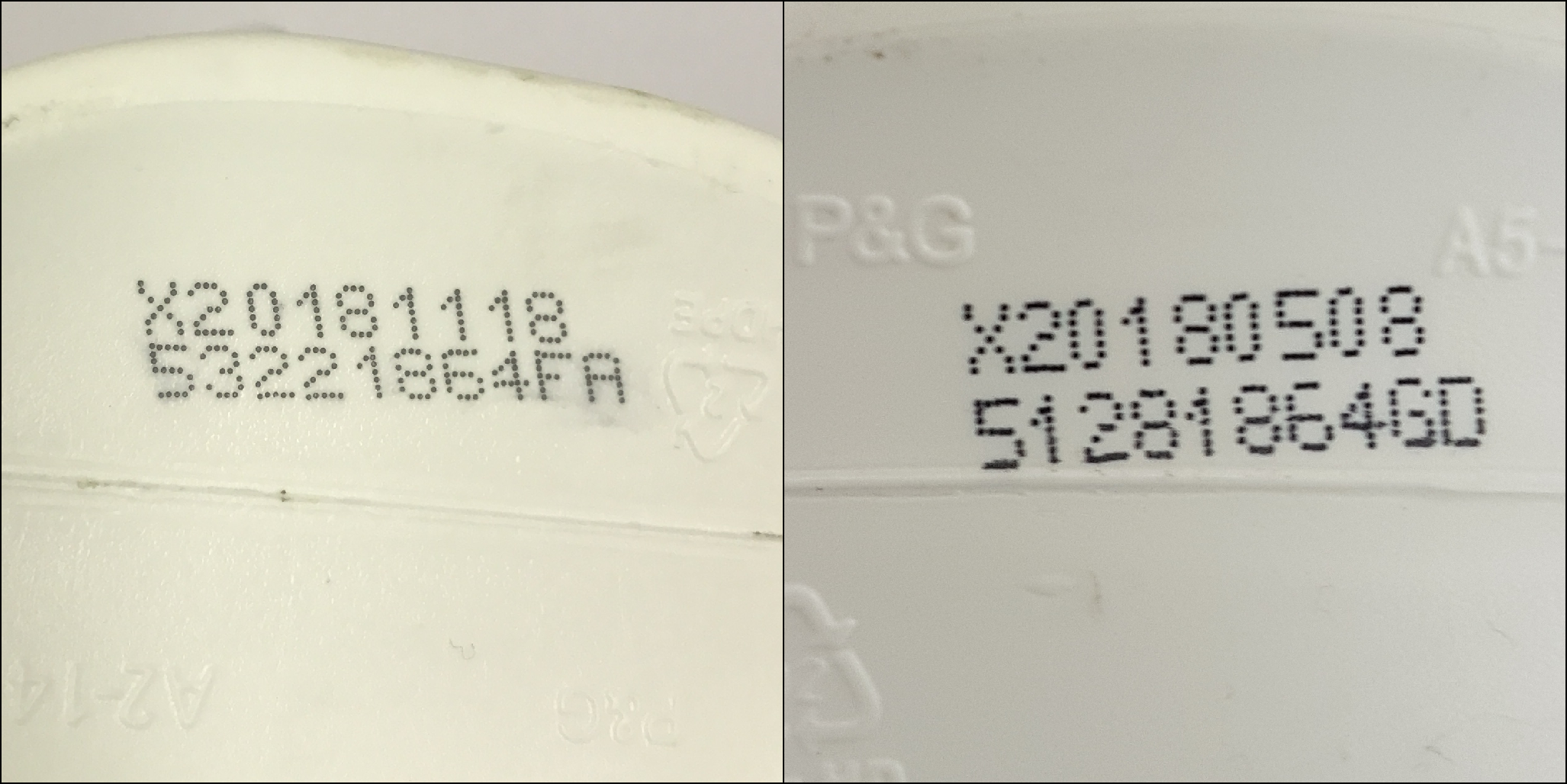}
    \caption{A pair of images from our dataset. The left was taken with an iPhone camera, while the right was taken with a Samsung.}
    \label{fig:batchcodes}
\end{figure}

Our dataset consists of $3090$ RGB images of the undersides of shampoo bottles. These images are of size $1024 \times 1024$ and are cropped tightly around the bottle's batch code, which is a two-line alphanumeric serial number printed by a dot matrix printer (see Figure~\ref{fig:batchcodes}). The crops are from roughly the same area of the original images, so they should cover mostly the same region of the cameras' sensors. This means they should contain roughly the same sensor pattern noise, up to some random translation. The batch codes of the two manufacturers' products are expected to differ in some potentially very subtle ways, hence the relatively high resolution of our images.

Our images are captured with five different cameras: iPhone, Huawei, Samsung, Redmi and Vivo. In the base dataset these cameras occur at equal frequencies among the two manufacturers, but by excluding certain combinations of camera and label from the dataset, we can introduce correlations between camera type and class label. Since we were aware of the camera bias issue at the time our dataset was collected, care was taken to remove all sources of domain bias (e.g. different people photographing bottles from each manufacturer, and perhaps holding the bottles differently). To this end, the images were acquired by four different people who each photographed an equal number of images from each manufacturer and with each camera, all in the same room, under controlled lighting conditions. This means that domain bias should only exist if we deliberately induce it by excluding certain manufacturer / camera combinations. It also means that background distractors should be uncorrelated with manufacturer and camera type. The test set is a random $10\%$ of the samples, on which the model is never trained.

\section{Experiments}
\label{sec:experiments}

To address the questions raised in Section~\ref{sec:intro}, we run a series of classification experiments on variations of our dataset with camera/label correlation artificially introduced. All experiments are trained to convergence with the Adam optimizer \cite{kingma2014adam}, using a learning rate of $10^{-4}$, a weight decay of $0$, and the categorical cross entropy loss function. All accuracy numbers we report are averaged over four runs with different random number generator seeds. We perform all our experiments with five commonly used CNN architectures, all of which are pretrained on ImageNet and fine-tuned on our tasks with no layer freezing.

\subsection{Camera Classification}

\begin{figure}
    \centering
    \includegraphics[width=0.8\linewidth]{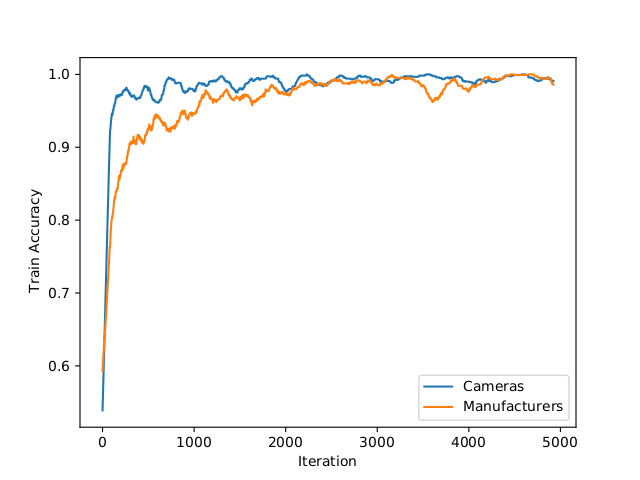}
    \caption{A pretrained ResNet34 model learns to recognize manufacturers very quickly, and learns to recognize cameras even faster.}
    \label{fig:camera_accuracy}
\end{figure}

As a basic sanity check, we verify here that state of the art vision models can very easily classify which camera took the image in this dataset. This corroborates the work of \cite{bondi2017preliminary} \cite{tuama2016camera} for our own datasets and cameras. Table~\ref{tab:camera-classification} shows that very high test accuracies can be achieved on this task across a range of architectures. As Figure~\ref{fig:camera_accuracy} shows, a pretrained ResNet34 not only achieves high accuracy at camera classification but does so very quickly. Camera recognition is learned faster than manufacturer recognition, suggesting that a model which can minimize its loss by recognizing manufacturers or by cheating by camera recognition will tend toward the latter, as it is somehow easier.

\begin{table}[h!]
    \centering
    \begin{tabular}{c c}
        \toprule
        \textbf{Model} & \textbf{Test Accuracy}\\
        \midrule \midrule
        ResNet34 & 0.999 \\
        ResNet101 & 0.913 \\
        InceptionV3 & 0.998 \\
        AlexNet & 0.945 \\
        VGG16 & 0.974 \\
        \bottomrule
    \end{tabular}
    \vskip 3pt
    \caption{Accuracy on the test set when classifying which camera took an image.}
    \label{tab:camera-classification}
\end{table}

\subsection{Manufacturer Classification}

We investigate our primary task, manufacturer classification, under three settings, which we refer to as Balanced, Partial, and Disjoint. In the Balanced setting, we use the full training set and there are no correlations between camera type and class label. In Partial, we use the same training set but with only iPhone and Samsung cameras included. In Disjoint, we introduce correlations between camera and class label by including only Manufacturer 1 images taken with iPhone or Samsung cameras, and Manufacturer 2 images taken with Huawei or Redmi cameras. Our test set is the same in all cases, balanced across camera types with no camera/label correlations. 

\begin{table}[h!]
    \centering
    \begin{tabular}{c c c c}
        \toprule
        \textbf{Model} & \textbf{Balanced} & \textbf{Partial} & \textbf{Disjoint}\\
        \midrule \midrule
        ResNet34 & 0.974 & 0.957 & 0.505 \\
        ResNet101 & 0.969 & 0.921 & 0.505 \\
        InceptionV3 & 0.973 & 0.940 & 0.518 \\
        AlexNet & 0.929 & 0.893 & 0.573 \\
        VGG16 & 0.979 & 0.945 & 0.556 \\
        \bottomrule
    \end{tabular}
    \vskip 3pt
    \caption{Manufacturer classification test set accuracy of five models with different training setups.}
    \label{tab:manufacturer-detection}
\end{table}

Table~\ref{tab:manufacturer-detection} shows the results of manufacturer classification experiments on these three datasets. It is immediately apparent that while respectable accuracy is achieved when training on the Balanced dataset, an accuracy drop of around $30\%$ occurs when training on Disjoint. In fact, Disjoint accuracy is close to $50\%$, hardly better than random guessing, which is entirely expected if the model were basing its classifications on camera types, each of which has an equal number of images of each class in the test set.

\begin{figure}
    \centering
    \includegraphics[width=0.8\linewidth]{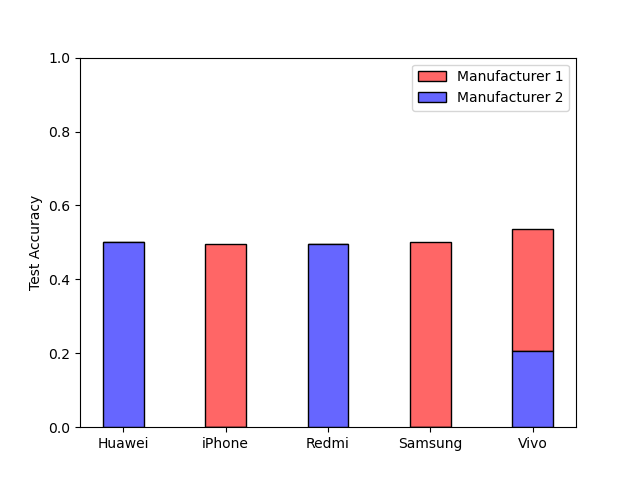}
    \caption{Test accuracy plot showing the distribution of predicted labels among correct outputs, for a ResNet34 trained on the Disjoint training set, in which all Manufacturer 1 images are iPhone or Samsung, and all Manufacturer 2 are Huawei or Redmi. For images from iPhone and Samsung cameras the model predicts only Manufacturer 1, while for Huawei and Redmi it predicts only Manufacturer 2, while for the unseen Vivo images it appears to guess randomly, achieving $54\%$ accuracy with a mostly even mix of both classes. Best viewed in color.}
    \label{fig:disjoint_manufacturers_resnet34}
\end{figure}

\begin{figure}
    \centering
    \includegraphics[width=0.8\linewidth]{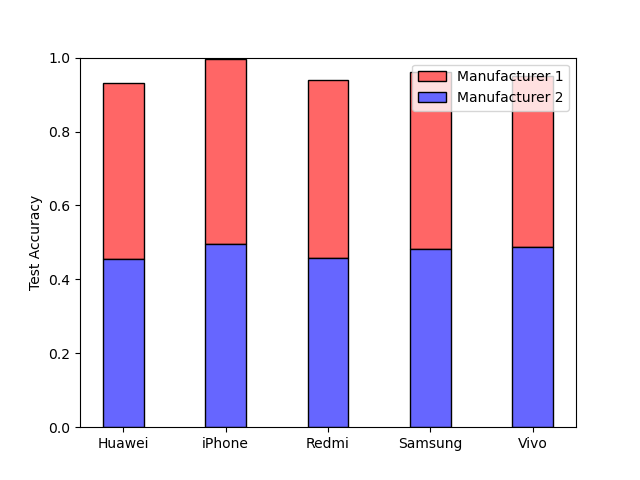}
    \caption{Test accuracy plot showing the distribution of predicted labels among correct outputs, for a ResNet34 trained on the Partial training set, in which camera type is uncorrelated with class label but only iPhone and Samsung images are present. Overall accuracy across all camera types is close to that achieved when trained on the full dataset, with little bias in favor of familiar camera types. This implies that in the absence of camera / label correlations, the model learns robust features for manufacturer classification, which generalize well to images from unseen cameras. Best viewed in color.}
    \label{fig:partial_manufacturers_resnet34}
\end{figure}

We can confirm that this drop in accuracy is due to camera bias by observing the model's behavior across camera types in the test set. As Figure~\ref{fig:disjoint_manufacturers_resnet34} shows, a ResNet34 trained on the Disjoint dataset predicts Manufacturer 1 exclusively on test images acquired by iPhone or Samsung cameras, and Manufacturer 2 overwhelmingly on Huawei and Redmi images. Similar behavior is observed in the other models.

\begin{figure*}
    \centering
    \includegraphics[width=\linewidth]{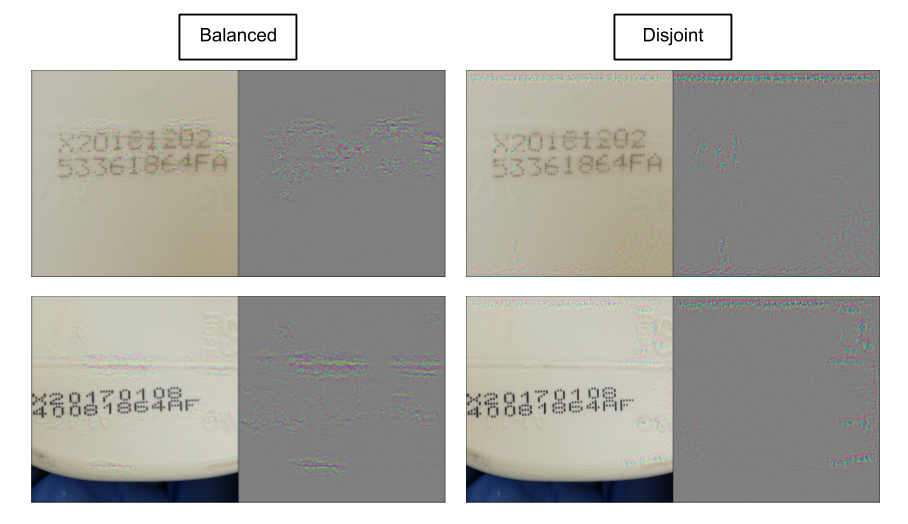}
    \caption{Adversarial perturbations applied to two images, classified by a ResNet34 model trained on the Balanced dataset (left) and the Disjoint dataset (right). The left image in each pair shows the input image with the perturbation amplified for visibility and overlaid on top, while the right image shows just the amplified perturbation itself. Strikingly different perturbations to the same image are observed depending on whether the model was trained without camera / label correlations (Balanced) or with them (Disjoint). Best viewed digitally, zoomed in.}
    \label{fig:adversarial_manufacturers}
\end{figure*}

It is interesting to note that AlexNet and VGG16 both score higher test accuracy after training on Disjoint than their more modern counterparts, ResNet34, ResNet101 and InceptionV3. One possible explanation for this is that AlexNet and VGG16 both use fully connected layers to produce their final output, while the more recent networks are fully convolutional, i.e. using a global average pooling layer to convert the final feature maps into a fixed size vector that is then classified by a single fully connected output layer. Fully connected layers have one parameter per input unit and hence require fixed size input, whereas convolutional layers can process arbitrary sized input by using the same convolutional weights at every location in the input. AlexNet and VGG16 therefore require input images to be downsampled to $224 \times 224$, whereas the fully convolutional networks receive the full $1024 \times 1024$ images. This suggests that camera identification exploits high frequency features (as opposed to geometric distortions caused by lens variations), which are partly destroyed during downsampling, thus preventing AlexNet and VGG16 from exploiting them.

When training on the Partial dataset, where only iPhone and Samsung images are present but no camera/label correlation exists, test accuracy is broadly similar to training on the full (Balanced) dataset. Not only is accuracy high, but as Figure~\ref{fig:partial_manufacturers_resnet34} shows, the model performs well on the unseen cameras. This implies that in the absence of camera/label correlation, the model learns a robust classification rule that is unaffected by camera type.

\subsection{Adversarial Attacks on Manufacturer Classifiers}

To gain some insight into the effect that camera / label correlations have on a trained model in terms of the patterns it learns to recognize, we perform adversarial attacks on trained models and visualize the perturbations that flip a trained model's judgement of an image from Manufacturer 2 to 1. Adversarial attacks are small perturbations to input images, imperceptible to the human eye, which nonetheless are sufficient to fool a model into classifying that image as whatever the attacker wishes \cite{szegedy2013intriguing}. They are easily generated by gradient ascent in image space, backpropagating the negative log likelihood of the target label into the image pixels and taking small steps in the direction of the resulting image gradient until the model's prediction favors our target (e.g. see Nguyen et a. \cite{nguyen2015deep}). By performing this process using the same image but different models and comparing the resulting image perturbations, we can learn something about how those models differ. 

Figure~\ref{fig:adversarial_manufacturers} shows that strikingly different adversarial perturbations are induced depending one which dataset the model was trained on. Perturbations that fool the Balanced model are focused around the batch code and other visible features of the bottle, such as the plastic seam, whereas those that fool the Disjoint model show a characteristic pink / green banding pattern in flat, featureless areas of the image. A distinct rainbow-like band of perturbation is also visible along the tops of images classified by the Disjoint model; these banding patterns at the tops and in featureless areas of images appear regardless of which input image the attack is performed on. 

Adversarial perturbations, when amplified for visibility, usually look like uninterpretable noise bearing little apparent resemblance to the target image class (e.g. Goodfellow et al. \cite{goodfellow2014explaining}), so it is interesting to see so much structure in our case. The appearance of banding patterns in flat regions provides some evidence against chromatic aberration, which should manifest at the edges of objects.

\subsection{Classification of Binary Masks}

\begin{figure}
    \centering
    \includegraphics[width=0.9\linewidth]{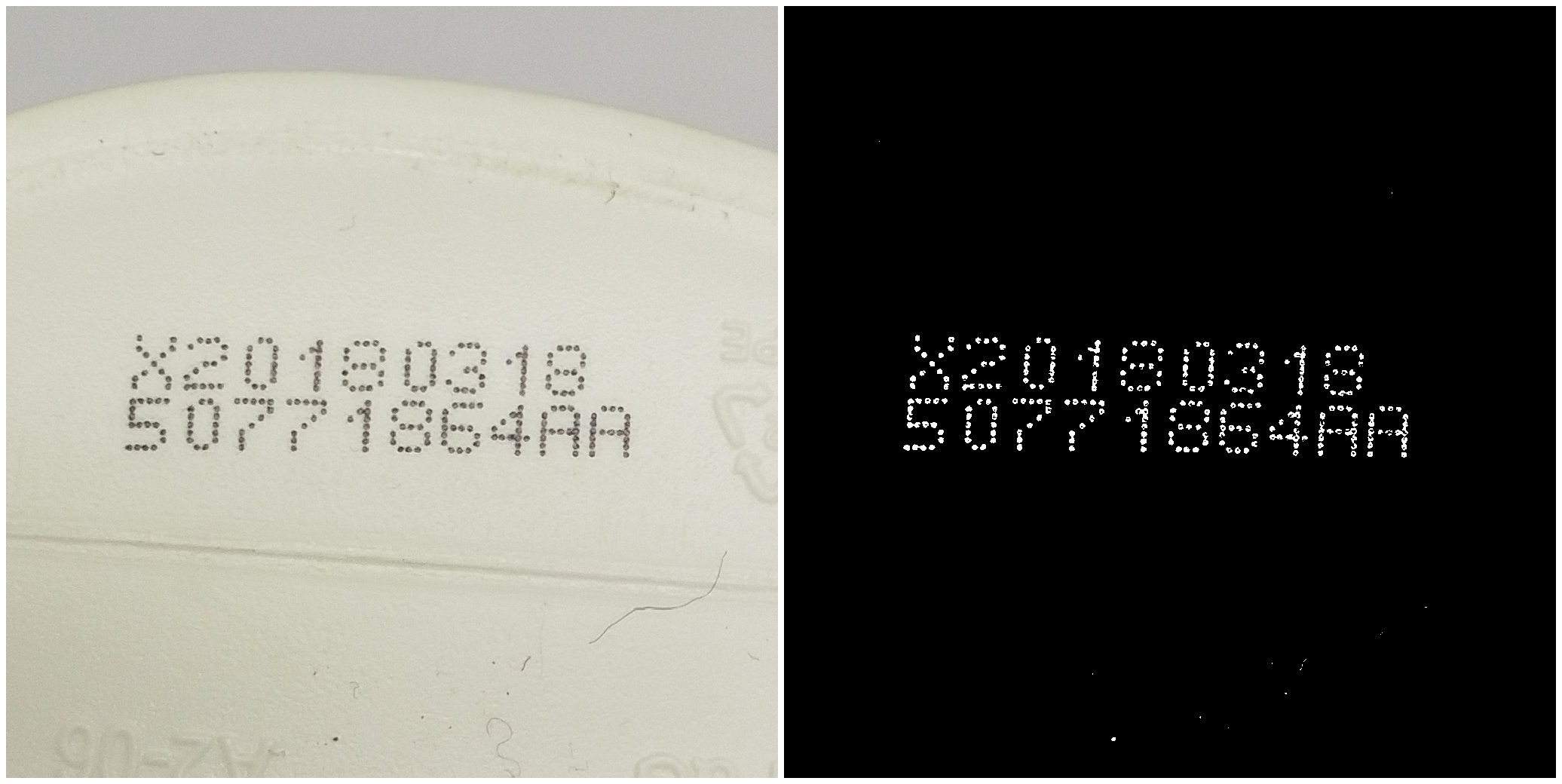}
    \caption{A bottle image with local mean thresholding applied, segmenting the batchcode dots. Origin camera classification does not work on such images, indicating that models use something other than the shape and position of the dots to classify cameras.}
    \label{fig:dotseg}
\end{figure}

As discussed in Section~\ref{sec:lit_review}, there is a finite set of image features that may be used to infer the camera from which an image originates. Since most of these features relate to color distribution or high frequency detail (i.e. patterns detectable within a small window), it seems likely that removal of these features would render camera identification impossible, and hence resolve the domain bias issues. To do this while preserving features that are likely relevant for robust manufacturer detection, we apply local mean thresholding to the images. This yields a binary image that effectively segments the dots of the batch codes while removing all elements of color and texture (see Figure~\ref{fig:dotseg}). Although lens distortion is typically more prevalent near the edges of images, 

As Table~\ref{tab:dotseg} shows, training on binary segmented images does not yield usable results on the manufacturer detection or camera classification tasks - in both cases, the test accuracy is close to the level expected of random guessing. As expected, we also observed no significant correlation between manufacturer classification test accuracy and camera type when training on the Disjoint dataset with binary thresholding. This largely rules out lens distortion or other large scale geometric artifacts as the source of camera bias, since these distortions would cause the dots to move and thus be visible in the binary thresholded images.

\newcommand\T{\rule{0pt}{2.9ex}}       
\newcommand\B{\rule[-1.2ex]{0pt}{0pt}} 

\begin{table}[t!]
	\centering
    \begin{tabular}{l c c}
        \toprule
        \multicolumn{1}{l}{\multirow{2}{*}{\T\B\T\textbf{Model}}} 
        & \multicolumn{2}{c}{\textbf{Test Accuracy}\B} \\ 
        \cline{2-3}

            & Manufacturers & Cameras \T\B \\ \hline \hline
            ResNet34 & 0.489 & 0.229\T \\
            ResNet101 & 0.530 & 0.186 \\   
            InceptionV3 & 0.525 & 0.270 \\
            Alexnet & 0.499 & 0.214 \\
            VGG16 & 0.616 & 0.384\B \\
        \bottomrule
    \end{tabular}
    \vskip 3pt
	\caption{Manufacturer and camera classification accuracy on the test set when trained (and tested) on binary segmented images (see Figure~\ref{fig:dotseg}).}
	\label{tab:dotseg}
\end{table}

\subsection{Classification of Color Jittered Images}
\label{sec:colorjitter}

\begin{figure}
    \centering
    \includegraphics[width=\linewidth]{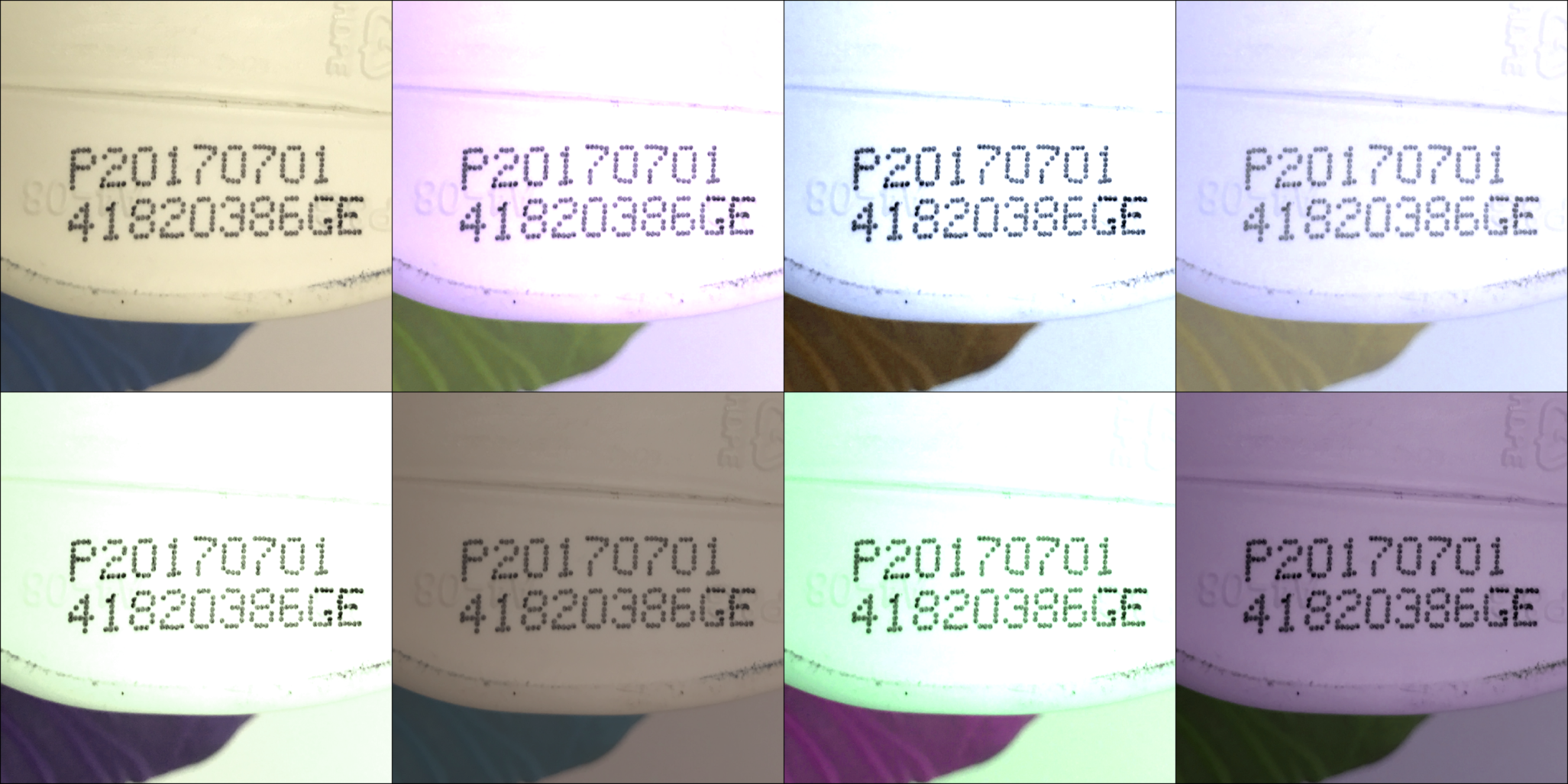}
    \caption{Color jitter augmentations applied to a single image (original in top left). Augmenting our training images with basic color distortions removes any correlations that may exist between class label and white balance, saturation, hue.}
    \label{fig:augmented}
\end{figure}

One hypothesis is that different cameras have subtly different color correction / white balancing settings, which a CNN could very easy detect and exploit, especially since the images were acquired in laboratory conditions with controlled lighting. We test this hypothesis by randomizing the hue, saturation, contrast and brightness of the images at training time, thus removing any correlation between camera type and global image color statistics. Table~\ref{tab:colorjitter} shows that even with the high level of color randomization used (see Figure~\ref{fig:augmented}), camera and manufacturer classification test accuracy remains high. Accuracy is somewhat diminished for the AlexNet and VGG16 architectures, which require downsampled images as input, suggesting that these features are still useful when high frequency features are less available. We also train models to recognize cameras from grayscale images, achieving test accuracies roughly identical to those in Table~\ref{tab:colorjitter}.

\begin{table}[t!]
    \centering
    \begin{tabular}{l c c}
        \toprule
        \multicolumn{1}{l}{\multirow{2}{*}{\T\B\T\textbf{Model}}} 
        & \multicolumn{2}{c}{\textbf{Test Accuracy}\B} \\ 
        \cline{2-3}
            & Manufacturers & Cameras \T\B \\ \hline \hline
            ResNet34 & 0.975 & 0.992\T \\
            ResNet101 & 0.961 & 0.995 \\   
            InceptionV3 & 0.974 & 0.998 \\
            Alexnet & 0.923 & 0.768 \\
            VGG16 & 0.972 & 0.883\B \\
        \bottomrule
    \end{tabular}
    \vskip 3pt
    \caption{manufacturer and camera classification test set accuracy when trained on images with randomized hue, saturation, contrast and brightness. Robust camera classification accuracy implies that image color statistics are not necessary for camera inference.}
    \label{tab:colorjitter}
\end{table}

\subsection{Classifying Cameras from Small Image Patches}

\begin{figure}
    \centering
    \includegraphics[width=0.5\linewidth]{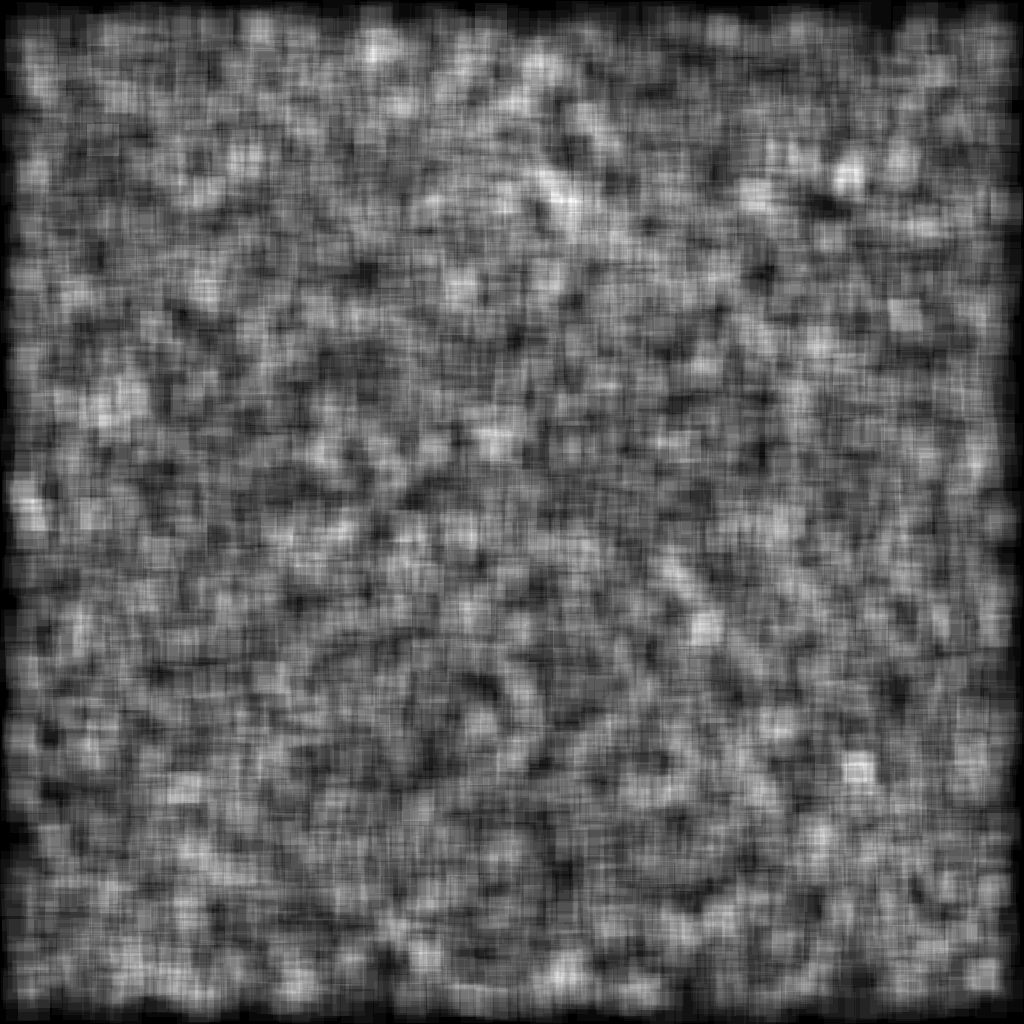}
    \caption{Heatmap representing relative classification accuracy of $32 \times 32$ crops at different locations in the image, averaged across images from the whole dataset. The lack of bias toward any particular part of the image implies that camera predictive patterns are present uniformly across the images.}
    \label{fig:heatmap}
\end{figure}

\begin{figure}
    \centering
    \includegraphics[width=\linewidth]{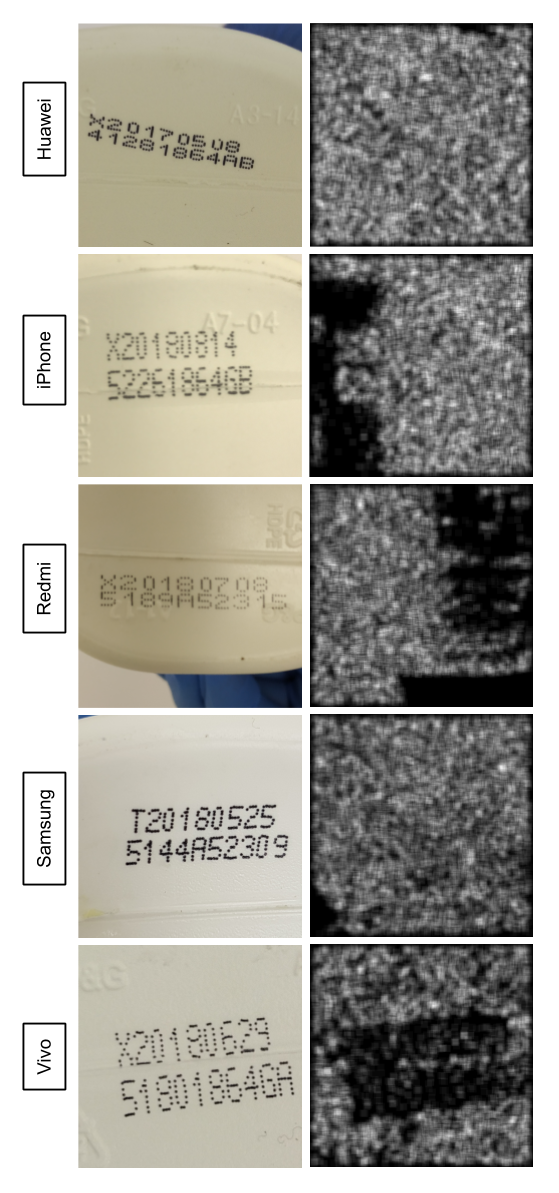}
    \caption{Heatmaps representing the camera identification accuracy on $32 \times 32$ patches at different locations in single images. An image from each camera is shown, and the predictions are all from the same ResNet34 checkpoint. The model is able to correctly classify patches from most locations on most images, but some significant dark patches occur.}
    \label{fig:singleimage_heatmap}
\end{figure}

With lens deformation and color statistics ruled out as camera identifying features, we turn our attention towards high frequency features. As discussed in Section~\ref{sec:lit_review}, such features could be introduced by various forms of fixed sensor pattern noise, dust particles stuck to the lens, and image processing / compression algorithms performed automatically by the camera. We investigate the role of high frequency features by training CNNs to classify cameras given only a random $32 \times 32$ crop of our original input images (upsampled to $224 \times 224$ for Alexnet and VGG16). As Table~\ref{tab:crop32} shows, camera identification accuracy remains surprisingly robust even when input is restricted to a $32 \times 32$ window. This strongly implies that high frequency features are sufficient for camera identification, and confirms that lens distortion is not required. However, it remains unclear whether these features are localized to certain regions of the image or present uniformly. Figure~\ref{fig:heatmap} shows an accuracy heatmap, constructed by repeatedly sampling $32 \times 32$ crops from our training set and drawing a white square at the location of each correctly classified crop. This shows that classification accuracy is independent of the location of the crop, at least when averaged over the whole dataset. This implies that whatever pattern is being exploited occurs uniformly across the images on average. Figure~\ref{fig:singleimage_heatmap} shows how classification accuracy for $32 \times 32$ crops varies across five individual images, one from each camera. 

\begin{table}[h!]
    \centering
    \begin{tabular}{c c}
        \toprule
        \textbf{Model} & \textbf{Test Accuracy}\\
        \midrule \midrule
        ResNet34 & 0.665 \\
        ResNet101 & 0.681 \\
        InceptionV3 & 0.948 \\
        AlexNet & 0.770 \\
        VGG16 & 0.872 \\
        \bottomrule
    \end{tabular}
    \vskip 3pt
    \caption{Camera classification test accuracy when trained only on random $32 \times 32$ crops of the input data.}
    \label{tab:crop32}
\end{table}

\vskip -30pt
\subsection{Generalizing from Left Field of View to Right}

Pixel non-uniformity (PNU) noise, as described in \cite{lukavs2006digital}, is a high frequency noise fingerprint, manifested as randomly varying sensitivities of individual sensor pixels to light. We would expect such a noise fingerprint to be non-repeating, that is, the noise pattern in one part of an image should be different to that in other parts. If the models are learning to recognize cameras by recognizing their PNU noise fingerprints, they should therefore be incapable of recognizing cameras from patches of noise fingerprint they have not encountered during training. We therefore test our models' reliance on PNU noise by training them on only the left halves of our Balanced training set images, and testing on the right halves. If they are reliant on PNU noise then generalization to the right halves of images should be poor. As Table~\ref{tab:leftside} shows, this is not the case, therefore PNU noise is unlikely to be the primary source of camera identifying information.

\begin{table}[h!]
    \centering
    \begin{tabular}{c c}
        \toprule
        \textbf{Model} & \textbf{Test Accuracy}\\
        \midrule \midrule
        ResNet34 & 0.992 \\
        ResNet101 & 0.889 \\
        InceptionV3 & 0.999 \\
        AlexNet & 0.818 \\
        VGG16 & 0.851 \\
        \bottomrule
    \end{tabular}
    \vskip 3pt
    \caption{Camera classification accuracy on right halves of images after training on the left halves. Strong generalization to an unseen area of the training images implies that PNU noise fingerprints of the sort discussed by Lukas et al. \cite{lukavs2006digital} are unlikely to be the mechanism by which CNNs are recognizing cameras, because the noise fingerprint on the right side of the images will be different to that on the left side.}
    \label{tab:leftside}
\end{table}

\section{Conclusion}
\label{sec:conclusion}

We have shown that CNNs learn to exploit camera / class label correlations in an image classification dataset in which such correlations are present. By recognizing the camera that acquired an image, CNNs are able to infer the class label without learning any features that are relevant to the task (in our case, manufacturer classification), as evidenced by poor generalization to images where the camera / label correlation is broken. This finding has relevance both to fine grained classification and to algorithmic transparency and accountability. We also show that CNNs are capable of learning to infer origin cameras when explicitly trained to do so, corroborating the results of Bondi et al. and Tuama et al. \cite{bondi2017preliminary, tuama2016camera}. We test these phenomena across five different CNN architectures and show that the effects are common to all of them, although lesser among AlexNet and VGG16, the two architectures whose inputs must be downsampled to a smaller size due to the use of fully connected layers. We have also performed a number of experiments to gain insight into how CNNs are recognizing cameras, the results of which require some discussion in this section. Section~\ref{sec:lit_review} outlines a number of potential sources of camera identifying information, and our experiments provide evidence for and against those hypotheses. 

A simple explanation for camera bias would be differences in average color statistics among cameras, caused by differences in white balance and color correction settings. This hypothesis is largely ruled out by the fact that CNNs still recognize cameras easily even when hue, saturation, contrast and brightness are randomized (see Figure~\ref{fig:augmented}, Table~\ref{tab:colorjitter}). Another potential explanation was lens distortion; if different cameras have different shaped lenses then there may be slight differences in geometric distortion (e.g. radial lens distortion \cite{choi2006source}). This hypothesis too is ruled out, by the fact that CNNs are incapable of inferring cameras from binary segmented images (see Figure~\ref{fig:dotseg}, Table~\ref{tab:dotseg}). Geometric distortions would be visible in the spacing of the dots from the batch codes, which are the only features visible in these images. Chromatic aberration is also unlikely since it should only be visible at the edges of objects, not in flat regions (Figure~\ref{fig:adversarial_manufacturers},\ref{fig:singleimage_heatmap}), and should also be undetectable in grayscale images (Section~\ref{sec:colorjitter}). These results increase the likelihood that texture, which is absent in segmented images but preserved in color randomized images, plays an important role. High camera recognition accuracy on $32 \times 32$ random crops (Table~\ref{tab:crop32}), including in empty patches of the image where it is hard to imagine what features besides faint, high frequency texture are available (Figure~\ref{fig:singleimage_heatmap}), increases this likelihood further.

There are two likely sources of camera correlated texture: pixel non-uniformity (PNU) noise, and the camera's on-board image processing, which typically includes algorithms such as kernel filtering, image sharpening and compression (both discussed by Lukas et al. \cite{lukavs2006digital}). PNU noise is dominated by a fixed multiplicative noise pattern that is introduced during manufacturing, as such we would expect different noise patterns in different parts of the field of view, as opposed to a repeating pattern. The fact that CNNs trained on the left hand sides of images generalize well to the right hand sides of those images (Table~\ref{tab:leftside}) therefore implies that PNU noise is not crucial for camera recognition, since they should not be able to recognize unseen noise patterns on the right side of the images. The fact that AlexNet and VGG16 are also able to recognize cameras from downsampled images is also strong evidence against PNU noise, which should be undetectable after downsampling.

By a process of elimination, the most likely explanation therefore seems to be on-camera image processing algorithms. We do not consider these results to be conclusive; a conclusive answer would require full knowledge of the original cameras, which we do not have. Further research is required to ascertain exactly which textural features are exploited by CNNs to recognize cameras.

{\small
\bibliographystyle{ieee_fullname}
\bibliography{refs}
}

\end{document}